\documentclass[lettersize,journal]{IEEEtran}
\usepackage{amsmath,amsfonts}
\usepackage{algorithmicx}
\usepackage{algorithm}
\usepackage{array}
\usepackage{textcomp}
\usepackage{stfloats}
\usepackage{xcolor}
\usepackage{tcolorbox}
\usepackage{url}
\usepackage{verbatim}
\usepackage{graphicx}
\usepackage{cite}
\usepackage{mathtools}
\usepackage{graphicx}
\usepackage{multirow}
\usepackage{booktabs}
\usepackage{array}
\usepackage{enumitem}
\usepackage{siunitx}
\usepackage{makecell}
\usepackage{amsmath}
\usepackage{pifont}
\usepackage{algpseudocode}
\newtheorem{definition}{Definition}
\newtheorem{theorem}{Theorem}
\usepackage{soul, xcolor}

\usepackage{graphicx}
\usepackage{caption}
\usepackage{subcaption} 

\usepackage{multirow}
\usepackage{booktabs}

\newcommand{\makepromptbox}[2]{
\begin{center}
\begin{tcolorbox}[colback=brown!5!white,
                  colframe=brown!50!black,
                  colbacktitle=brown!75!black,
                  width=0.48\textwidth,
                  title=#1]
\normalsize #2
\end{tcolorbox}
\end{center}
}

\begin{document}

\title{Causal Prompting for Implicit Sentiment Analysis with Large Language Models}

\author{Jing Ren,~\IEEEmembership{Student Member, IEEE}, Wenhao Zhou, Bowen Li, Mujie Liu,~\IEEEmembership{Student Member, IEEE}, Nguyen Linh Dan Le,~\IEEEmembership{Student Member, IEEE}, Jiade Cen, Liping Chen,~\IEEEmembership{Student Member, IEEE}, Ziqi Xu, Xiwei Xu, and Xiaodong Li,~\IEEEmembership{Fellow,~IEEE}
\thanks{Jing Ren, Wenhao Zhou, Bowen Li, Nguyen Linh Dan Le, Jiade Cen, Liping Chen, Ziqi Xu, and Xiaodong Li are with the School of Computing Technologies, RMIT University, Melbourne 3000, Australia. (emails: jing.ren@ieee.org; wenhaozhou1112@gmail.com; bowen.li@rmit.edu.au; dan.le@ieee.org; jiade.cen@outlook.com; lp.chen@ieee.org; ziqi.xu@rmit.edu.au; xiaodong.li@rmit.edu.au)}
\thanks{Mujie Liu is with the Institute of Innovation, Science and Sustainability, Federation University Australia, Ballarat 3353, Australia.(email: mujie.liu@ieee.org)}
\thanks{Xiwei Xu is with CSIRO's Data61, Australia. (email: xiwei.xu@data61.csiro.au)}
\thanks{Corresponding authors: Ziqi Xu, Xiwei Xu}
}



\maketitle

\begin{abstract}
Implicit Sentiment Analysis (ISA) aims to infer sentiment that is implied rather than explicitly stated, requiring models to perform deeper reasoning over subtle contextual cues. While recent prompting-based methods using Large Language Models (LLMs) have shown promise in ISA, they often rely on majority voting over chain-of-thought (CoT) reasoning paths without evaluating their causal validity, making them susceptible to internal biases and spurious correlations. To address this challenge, we propose CAPITAL, a causal prompting framework that incorporates front-door adjustment into CoT reasoning. CAPITAL decomposes the overall causal effect into two components: the influence of the input prompt on the reasoning chains, and the impact of those chains on the final output. These components are estimated using encoder-based clustering and the NWGM approximation, with a contrastive learning objective used to better align the encoder's representation with the LLM's reasoning space. Experiments on benchmark ISA datasets with three LLMs demonstrate that CAPITAL consistently outperforms strong prompting baselines in both accuracy and robustness, particularly under adversarial conditions. This work offers a principled approach to integrating causal inference into LLM prompting and highlights its benefits for bias-aware sentiment reasoning. The source code and case study are available at: \url{https://github.com/whZ62/CAPITAL}.
\end{abstract}

\begin{IEEEkeywords}
Causal inference, large language models, implicit sentiment analysis
\end{IEEEkeywords}

\section{Introduction}
\begin{figure}[t]
    \centering
    \includegraphics[width=0.45\textwidth]{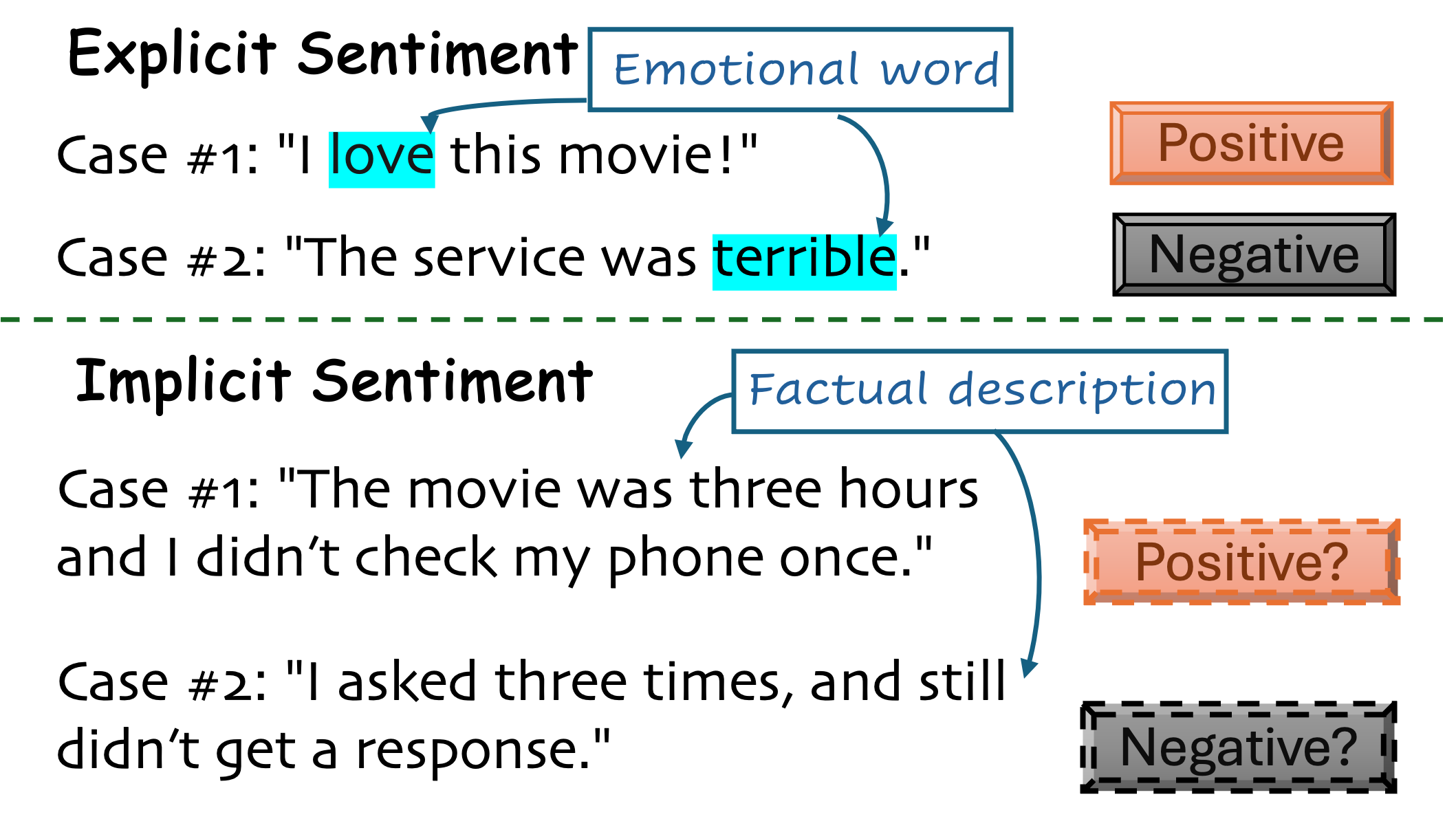}
    \caption{Examples comparing explicit and implicit sentiment analysis. Sentences used in ESA contain emotional words, while those used in ISA convey sentiment through factual descriptions.}
    \label{fig:figure1}
\end{figure}

\IEEEPARstart{S}{entiment} analysis (SA) refers to the computational task of detecting and interpreting emotions, opinions, or attitudes expressed in text. It is widely utilised in applications such as social media monitoring and customer feedback analysis. A common variant of this task is \textit{explicit sentiment analysis} (ESA), which focuses on detecting sentiment conveyed through clearly emotional or opinionated words, such as excellent, terrible, or disappointed. This task is typically approached using sentiment lexicons or supervised models trained on labelled data containing overt sentiment cues~\cite{yu2022exploring}. However, in many real-world scenarios, sentiment is expressed implicitly rather than through overt emotional language, which limits the applicability of ESA and highlights the importance of implicit sentiment analysis (ISA). ISA aims to identify sentiment that is not directly stated but inferred from subtle semantic cues and contextual information, as illustrated in Figure~\ref{fig:figure1}. It requires a deeper understanding of discourse, background knowledge, and implied meaning, making it significantly more challenging. For example, a sentence like ``The wait time was endless'' may be misclassified as neutral due to the absence of explicitly negative words, even though it clearly expresses dissatisfaction. This limitation arises because many existing methods rely primarily on surface-level lexical features and fail to capture the layered reasoning that ISA demands. As a result, successfully addressing ISA calls for models capable of advanced cognitive functions such as commonsense reasoning, multi-hop inference, and pragmatic understanding~\cite{fei2023reasoning}.

In recent years, the popularity of Large Language Models (LLMs) and their advanced reasoning capabilities make them a powerful tool for tackling the complexities of ISA, offering scalable and adaptable solutions. For example, Fei et al.~\cite{fei2023reasoning} propose a framework, THOR, which uses chain-of-thought self-consistency (CoT-SC) reasoning and commonsense integration to infer implicit emotions by analysing nuanced language and intent. Specifically, THOR reinforces the reasoning process by selecting the most frequent answer from a series of customised chains of thought. This framework employs LLMs under both prompting and fine-tuning settings; however, while fine-tuning is feasible, it is often costly, limited to open-source LLMs, and tends to suffer from poor generalizability. In contrast, prompting enables the model to perform effectively in zero-shot or few-shot settings with minimal resource requirements. Nevertheless, recent studies find that LLMs exhibit internal biases that lead to spurious correlations with the query, thereby limiting the model’s ability to leverage contextual knowledge for accurate polarity prediction. Although THOR shows effectiveness in reducing random reasoning errors through the self-consistency strategy, it does not adequately mitigate the spurious associations introduced by these internal biases within the LLM.

\begin{figure}[t]
	\centering
	\begin{subfigure}[b]{0.25\linewidth}
		\centering
		\includegraphics[width=\linewidth]{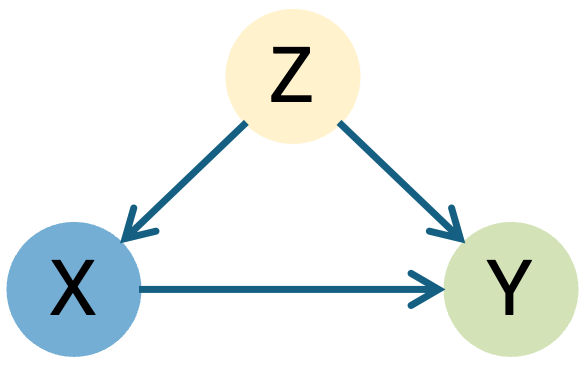}
		\caption{}
		\label{fig:fig2a}
	\end{subfigure}
	\hfill
	\begin{subfigure}[b]{0.33\linewidth}
		\centering
		\includegraphics[width=\linewidth]{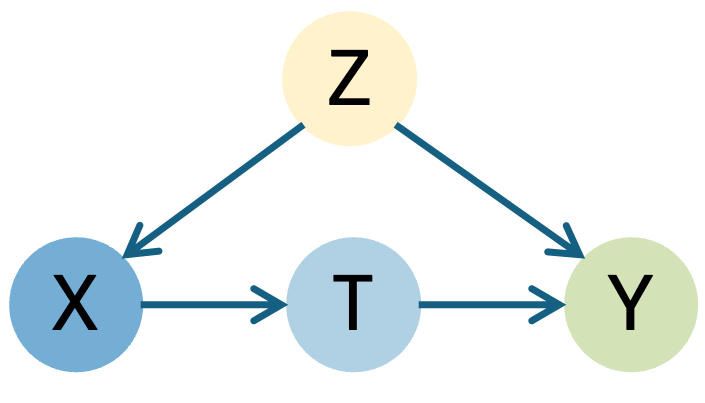}
		\caption{}
		\label{fig:fig2b}
	\end{subfigure}
	\hfill
	\begin{subfigure}[b]{0.33\linewidth}
		\centering
		\includegraphics[width=\linewidth]{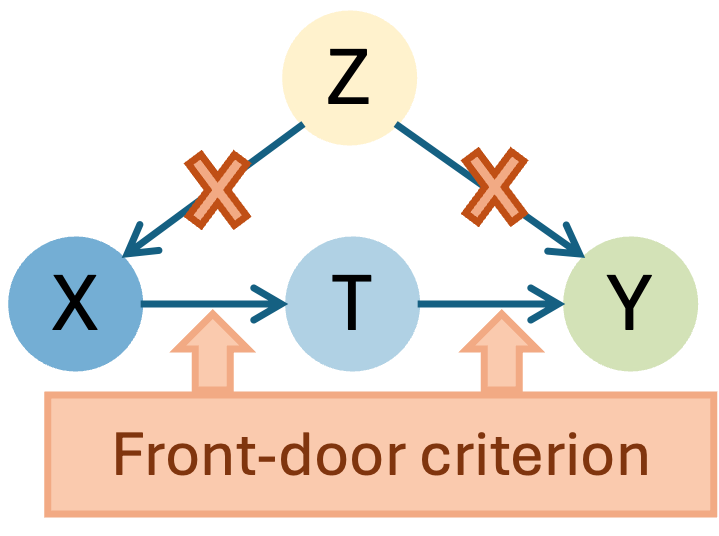}
		\caption{}
		\label{fig:fig2c}
	\end{subfigure}
	\caption{Three structural causal models (SCMs) representing reasoning in LLMs: (a) general reasoning without CoTs; (b) CoT is conceptualised as a mediator; (c) applying the front-door criterion to the CoT reasoning process to mitigate the impact of confounders. Here, $X$ is the query, $Y$ is the answer, $T$ is the CoT, and $Z$ is the latent confounder.}
	\label{fig:fig2}
\end{figure}

\begin{figure*}[t]
    \centering
    \includegraphics[width=0.96\textwidth]{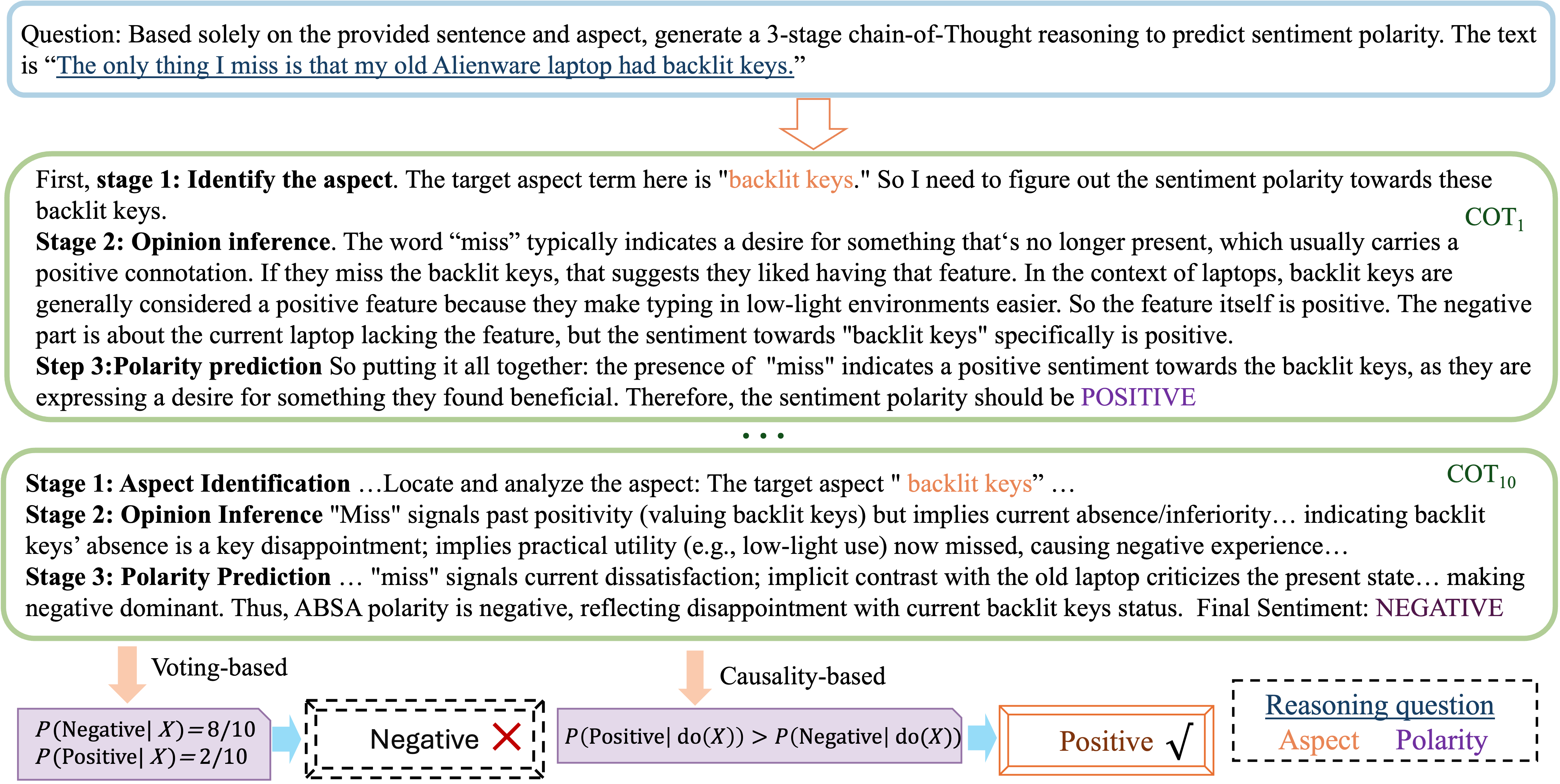}
    \caption{An example illustrating internal bias in LLMs. Voting-based methods such as THOR~\cite{fei2023reasoning} select the most frequent answer among all generated CoTs, which may lead to incorrect predictions. In contrast, our causality-based method selects the answer with the highest estimated causal effect, resulting in a correct prediction.}
    \label{fig:cot}
\end{figure*}

Existing studies show that causal reasoning can be incorporated into the prompting process to assess the causal effect of the query or the CoT on the final answer, thereby identifying more dependable reasoning pathways~\cite{DBLP2403-02738,DBLP:conf/acl/Wu0CWRKRM24}. Figure~\ref{fig:cot} presents an illustrative example comparing THOR with our framework. Unlike THOR, which adopts a self-consistency strategy by selecting the most frequently generated answer, our framework identifies the correct answer by estimating the causal effect of the query on each candidate output. The answer associated with a higher estimated causal effect is regarded as more trustworthy. This causal perspective allows our method to better distinguish between correct and spurious reasoning paths, resulting in more accurate predictions.

Furthermore, we formally categorise existing methods for ISA from a causal perspective. Figure~\ref{fig:fig2a} represents conventional approaches, including both traditional methods and LLM-based methods that do not incorporate CoTs, where the model maps the input query directly to an answer based on surface-level features. In contrast, Figure~\ref{fig:fig2b} depicts CoT-based reasoning in LLMs (e.g., THOR), which introduces intermediate reasoning steps to enhance answer generation. Despite these differences, both structural causal models reveal a common limitation: the presence of a latent confounder \( Z \), such as hidden knowledge or implicit bias, which simultaneously influences both the query and the answer. As a result, neither method is capable of producing unbiased predictions. This observation highlights the necessity of a causal prompting framework capable of estimating the causal effect of the query on the answer in a bias-free manner, thereby improving ISA performance.

In this paper, we propose a novel CAusal Prompting framework for Implicit sentimenT AnaLysis (CAPITAL), which leverages front-door adjustment to mitigate internal bias (as shown in Figure~\ref{fig:fig2c}) and produce more trustworthy polarity predictions. As the theoretical foundation of our framework, front-door adjustment is a principled method in causal inference that enables the reduction of bias introduced by latent confounders. To address the variability in answers generated from multiple CoTs, CAPITAL ranks the outputs based on their estimated causal effects, thereby selecting the most reliable answer.

The contributions of this paper are summarised as follows:
\begin{itemize}
    \item We present a causal perspective on implicit sentiment analysis using structural causal models, providing a theoretical foundation for de-biasing polarity predictions in LLMs.
    \item We propose CAPITAL, a novel causal prompting framework based on front-door adjustment, which supports both open-source and closed-source LLMs and ranks candidate answers by estimating their causal effects.
    \item We conduct extensive experiments on multiple implicit sentiment benchmarks, showing that CAPITAL consistently outperforms state-of-the-art prompting baselines in both accuracy and robustness.
\end{itemize}

The remainder of this paper is structured as follows. Section~\ref{sec:related} reviews related work. Section~\ref{sec:method} introduces the CAPITAL framework and elaborates on the proposed causal prompting methodology. Section~\ref{sec:experiments} presents the experimental setup and results, including comparisons with baselines, robustness analysis, and ablation studies. Finally, Section~\ref{sec:conclusion} concludes the paper and outlines directions for future work.

\section{Related Work}
\label{sec:related}
\subsection{Implicit Sentiment Analysis}
ISA focuses on detecting sentiments that are not explicitly expressed in text but are implied through context, common sense, or pragmatics~\cite{tubishat2018implicit}. Early approaches rely on sentiment lexicons and handcrafted rules. Subsequent deep learning methods adopt CNNs~\cite{chen2015convolutional}, RNNs~\cite{tang2015rnn}, and Transformers~\cite{devlin2019bert} to automatically learn features from data. Attention mechanisms further enhance performance by identifying sentiment-relevant context in sentences. More recent methods leverage pre-trained models to align implicit and explicit sentiment representations through data augmentation~\cite{wang2022contrastive}, contrastive learning~\cite{li2021learning}, and graph-based reasoning~\cite{wang2020relational}. However, these models often require large labelled datasets and struggle to generalise or explain complex reasoning steps.

LLMs have shown strong capabilities for ISA due to their advanced language understanding~\cite{zhang2024sentiment}. Emotion-aware generation~\cite{ouyang2024aspect} and prompt-based methods such as THOR~\cite{fei2023reasoning} and RIVSA~\cite{lai2025rvisa} further improve performance by enhancing CoT reasoning. Nevertheless, the reasoning patterns of these methods typically follow the causal structures illustrated in Figure~\ref{fig:fig2a} or~\ref{fig:fig2b}, which makes them vulnerable to spurious correlations introduced by latent confounders~\cite{zhang2025causal}. As a result, they may fail to capture the true causal relationships between input and sentiment, leading to biased predictions. To address this issue, we introduce a front-door adjustment mechanism that explicitly models the causal pathways underlying CoT reasoning, aiming to reduce confounding bias and improve robustness in ISA tasks, as illustrated in Figure \ref{fig:fig2c}.

\subsection{Prompting with LLMs}
With the rise of LLMs, prompting has become a widely used paradigm for adapting models to new tasks without fine-tuning~\cite{debnath2025comprehensive}. Basic prompting strategies, such as zero-shot and few-shot learning~\cite{brown2020language}, guide the model using either direct queries or a few in-context examples. Given LLMs' sensitivity to prompt phrasing, researchers have explored both handcrafted templates~\cite{reynolds2021prompt} and automated prompt optimisation methods~\cite{li2025survey} to improve performance. To address complex reasoning tasks, CoT prompting was proposed to elicit intermediate reasoning steps~\cite{wei2022chain}, while self-consistency further enhanced robustness by sampling multiple reasoning paths and selecting the most frequent output~\cite{wangself}. Building on these techniques, THOR~\cite{fei2023reasoning} integrates CoT and SC prompting into ISA, guiding the LLM to perform step-by-step inference and improving its ability to detect subtle emotional cues and implicit intent. However, such approaches typically rely on naïve voting mechanisms to select the final answer, without assessing the quality or causal validity of individual reasoning paths. As a result, they remain vulnerable to internal biases and spurious correlations in the generated outputs.

\subsection{Causal Inference for LLMs}
Causal inference aims to uncover the underlying causal relationships between variables using principled scientific methods~\cite{pearl2016causal}. With strong theoretical guarantees, various methods have been developed to estimate causal effects even in the presence of unobserved confounders~\cite{10791303,DBLP:conf/iclr/XuCLL0Y24,hengXL0LL24,ChengXLLLGL24,ChengXLLLL23,DBLP:conf/pkdd/ChengXLLLL23}. These methods have been applied to numerous NLP tasks such as de-biasing~\cite{DBLP:conf/acl/Wu0CWRKRM24}, fake news detection~\cite{wang2022causal}, and question answering~\cite{bondarenko2022causalqa}. Most of these methods build on Pearl's causal theory to estimate causal effects from observational data~\cite{pearl2016causal}.

More recently, research has begun to integrate causal reasoning into prompting. For example, Causal Prompting~\cite{zhang2025causal} designs prompts aligned with assumed causal structures to encourage causally consistent responses. DeCoT~\cite{DBLP:conf/acl/Wu0CWRKRM24} embeds causal structures into CoT reasoning by applying front-door adjustment and leveraging instrumental variables to filter misleading reasoning paths induced by latent confounders. However, these methods are primarily tailored to general reasoning or knowledge-intensive tasks, and they do not define a customised prompting strategy for ISA. In contrast, we propose CAPITAL, a front-door causal prompting framework that estimates the causal effect of each candidate output without relying on confounders or external variables, enabling more robust and interpretable reasoning for ISA.

\section{Preliminaries}
In this section, we introduce the theoretical foundations of our framework by reviewing structural causal models and presenting the back-door and front-door adjustment criteria that underpin our causal formulation for ISA with LLMs. Due to page constraints, readers are referred to~\cite{pearl2009causality} for several foundational concepts in causality, including directed acyclic graphs (DAGs), the Markov condition, faithfulness, \textit{d}-separation, and \textit{d}-connection.

\subsection{Structural Causal Model}
In a SCM~\cite{pearl2016causal}, the causal relationships between variables are formalised by a directed acyclic graph (DAG) $\mathcal{G}=(\mathcal{V},\mathcal{E})$, where $\mathcal{V}$ and $\mathcal{E}$ denote the sets of nodes and directed edges, respectively. Here, an edge from $\mathcal{V}_i$ to $\mathcal{V}_j$ indicates that $\mathcal{V}_i$ is a direct cause of $\mathcal{V}_j$. We use a sequence of nodes $(\mathcal{V}_1,\mathcal{V}_2,\ldots,\mathcal{V}_n)$ to represent a path $\pi$ between $\mathcal{V}_1$ and $\mathcal{V}_n$, where each consecutive pair $(\mathcal{V}_i,\mathcal{V}_{i+1})$ is adjacent along the path. 

In ISA, we provide an input query $X$ to the LLM and receive the response $Y$ representing the sentiment polarity predicted by LLM. As shown in Figure~\ref{fig:fig2a}, for traditional LLM-based ISA methods without any reasoning process, the response depends solely on the query. However, during pre-training, LLMs may absorb spurious correlations between superficial patterns and output distributions. These correlations, often originated from large-scale web data, can induce implicit biases that adversely affect the model’s reasoning in ISA~\cite{aghajanyan2021intrinsic, ding2023parameter}. In this case, the causal effect of $X$ on $Y$ cannot be unbiasedly estimated.

\begin{figure*}[t]
    \centering
    \includegraphics[width=0.98\textwidth]{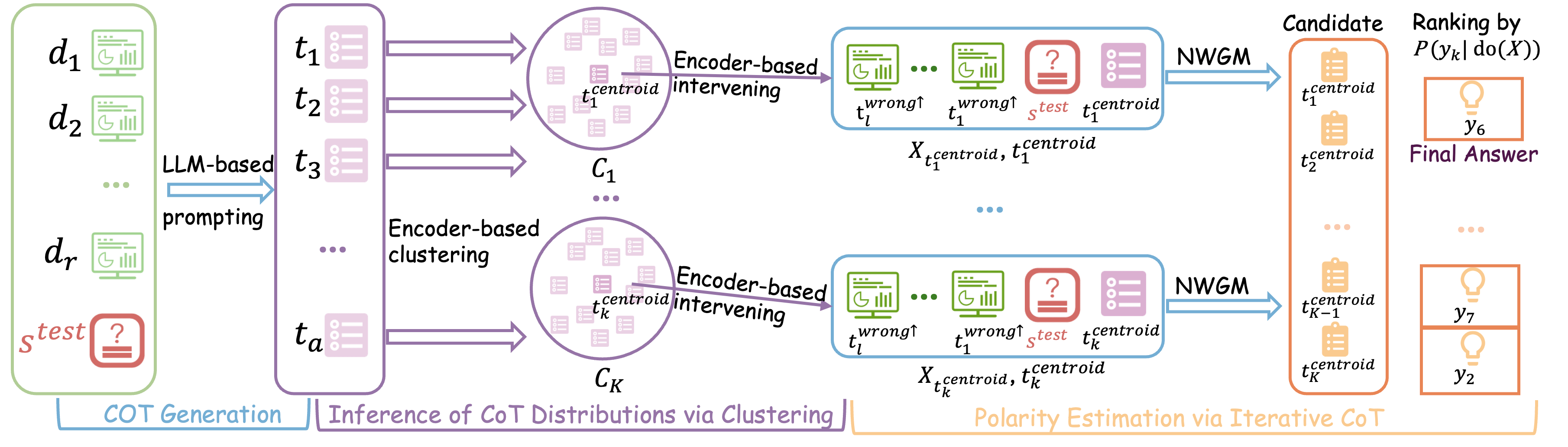}
    \caption{The overall framework of CAPITAL, which includes three stages: CoT generation, distribution inference via clustering, and polarity estimation. Given a prompt $x$ with $R$ in-context demonstrations and a test sentence $s^{\text{test}}$, the LLM generates $A$ distinct CoTs. These CoTs are encoded and clustered into $K$ groups using an encoder-based K-means algorithm, from which representative CoTs $t_k^{\text{centroid}}$ are selected. For each $t_k^{\text{centroid}}$, relevant demonstrations are retrieved to form a revised prompt $x_{t_k^{\text{centroid}}}$ via an NWGM-based selection mechanism. The LLM is then queried to obtain output predictions, and the final polarity is determined by selecting the answer associated with the highest estimated causal effect via the front-door adjustment formula.}
    \label{fig:framework}
\end{figure*}

\subsection{Back-door Adjustment}

The internal bias caused by these spurious correlations is formalised by an unobservable variable $Z$. As shown in Figures~\ref{fig:fig2a} and~\ref{fig:fig2b}, there is a structural dependency from $X$ to $Y$, but their relationship is confounded by this unobservable variable $Z$, forming two backdoor paths $X \gets Z \to Y$, where $X$ is no longer the sole direct cause of $Y$. To correctly estimate the causal effect of $X$ on $Y$, it is therefore necessary to block these backdoor paths. We define the backdoor criterion as follows:

\begin{definition}[Back-Door Criterion~\cite{pearl2009causality}]
    \label{def:BD}
    A set of variables $W$ satisfies the back-door criterion if
\begin{itemize}
    \item $W$ blocks all back-door paths from $X$ to $Y$.
    \item $W$ does not include any descendants of $X$.
\end{itemize}
\end{definition}

\begin{theorem}[Back-Door Adjustment~\cite{pearl2009causality}]
    If $W$ satisfies the back-door criterion relative to $X$ and $Y$, then the causal effect of $X$ on $Y$ is identifiable and is given by the following back-door formula:
	\begin{equation}
    \centering
    \label{eq:backdoor}
    P(Y \mid \mathrm{do}(X)) = \sum_{W} P(Y \mid X, W) P(W).
\end{equation}
\end{theorem}

Here, a {backdoor path} from $X$ to $Y$ is any path that starts with an arrow pointing into $X$ and leads to $Y$. The $\mathrm{do}(X)$ operator denotes an intervention setting $X$ to value $x$, effectively breaking all incoming edges to $X$ and isolating the causal effect of $X$ on $Y$. Therefore, the back-door adjustment allows us to estimate the causal effect by conditioning on an appropriate set $W$ that blocks confounding paths.

However, the variable $Z$ in Figures~\ref{fig:fig2a} and~\ref{fig:fig2b} satisfies the back-door criterion but remains unobserved, making the back-door adjustment inapplicable. Therefore, methods such as front-door adjustment are required to obtain an unbiased estimate of the causal effect.

\subsection{Front-door Adjustment}
Front-door adjustment is a widely used method to mitigate the impact of unobserved confounders~\cite{pearl2009causality}. Unlike the back-door criterion, which requires blocking all backdoor paths, the front-door criterion identifies the causal effect through an appropriate front-door adjustment variable, even in the presence of unobserved confounders. We introduce the formal definition as follows:

\begin{definition}[Front-Door Criterion~\cite{pearl2009causality}]
    \label{def:FD}
    A variable set $W$ satisfies the front-door criterion relative to the pair $(X, Y)$ if the following conditions hold:
\begin{itemize}
    \item $W$ fully mediates the effect of $X$ on $Y$;
    \item No unblocked back-door paths from $X$ to $W$;
    \item No unblocked back-door paths from $M$ to $Y$.
\end{itemize}
\end{definition}

\begin{theorem}[Front-Door Adjustment~\cite{pearl2016causal}]
    If $W$ satisfies the front-door criterion relative to $X$ and $Y$, then the causal effect of $X$ on $Y$ is identifiable and is given by the following front-door formula:
	\begin{equation}
		\label{eqa:FD2009}
		\begin{aligned}
			P(Y\mid \mathrm{do}(X)) = \sum_{W,X'}^{}P(Y\mid X',W)P(W\mid X)P(X'),
		\end{aligned}
	\end{equation}where $X'$ is a distinct realisation of treatment.
\end{theorem}

The aforementioned front-door criterion provides a theoretical foundation for identifying causal effects in the presence of unobserved confounders for binary intervention. However, in LLM-related tasks, the query variable $X$ is fixed and cannot be manipulated like a binary treatment as assumed in the original front-door adjustment formula. This limitation renders the standard front-door criterion inapplicable to ISA with LLMs. To address this challenge, we introduce a variant of the front-door adjustment formula tailored specifically for ISA scenarios involving LLMs in the following section.

\section{Methodology}
\label{sec:method}
In this section, we first outline the problem setting and introduce the notations used throughout the paper. Next, we describe the generation process of CoT reasoning. Subsequently, we present our proposed framework, CAPITAL. The overall architecture of CAPITAL is illustrated in Figure~\ref{fig:framework}.

\subsection{Problem Statement}
We consider the ISA task, where the goal is to identify the sentiment polarity toward a specific target term $g \subset S$ within a sentence $S$. The polarity label $Y$ is classified as positive, neutral, or negative. To address this task, we adopt a prompt-based framework leveraging an off-the-shelf LLM. The input to the LLM follows the prompt template as shown below:

\makepromptbox{}{Given the sentence $S$, what is the sentiment polarity towards $g$?}

Specifically, the LLM is prompted with the input query $X$ and generates CoT reasoning paths $T$ to support its inference. Based on the generated reasoning, the LLM produces a final sentiment prediction $Y$ for the target $g$. In the no-causality prompting framework, the LLM  predicts the sentiment label based on the input prompt, following:
\begin{equation}
    \hat{Y} = \arg\max_{Y}~P(Y \mid X)
\end{equation}

However, when an unobserved variable $Z$ exists, as shown in Figure~\ref{fig:fig2c}, the estimation of $P(Y \mid X)$ may be biased, since $X$ and $Y$ can be spuriously correlated through $Z$. In this case, $P(Y \mid X)$ does not reflect the causal effect of $X$ on $Y$, but rather captures associations that may be confounded. This leads to unreliable predictions, especially in tasks where the reasoning path should be grounded in causal mechanisms rather than surface-level correlations. To address this challenge, we obtain an unbiased estimate of the interventional distribution $P(Y \mid \mathrm{do}(X))$. By recovering this causal effect, we can rank candidate sentiment polarity predictions and select the one with the highest causal effect from the query, which we regard as the most reliable or correct sentiment classification.

\subsection{Generating CoT Reasoning Paths}
Inspired by~\cite{fei2023reasoning}, we adopt a similar multi-step reasoning framework for sentiment inference. Unlike existing LLM-based methods that directly generate the final prediction, our proposed CAPITAL framework decomposes the task into a structured three-step reasoning process, which first infers the latent aspect and opinion information before arriving at the final sentiment polarity.

Although our framework resembles THOR in terms of multi-step prompting, we do not incorporate self-consistency (i.e., voting) at each intermediate step, as it introduces substantial token overhead and may limit scalability. Instead, we treat the entire multi-step inference as a single reasoning trajectory, and subsequently apply a causality-based selection strategy to identify the most reliable final prediction. The three-step prompts are then constructed as follows.

\makepromptbox{}{
Based solely on the provided sentence and aspect, generate a 3-stage Chain-of-Thought reasoning to predict sentiment polarity.\newline

Reason through the text step by step and provide the final answer in the end. I will provide a reasoning process, and please improve the reasoning process and make sure you get the correct answer.

\medskip
\textbf{Reasoning Structure:}

{Stage 1: Aspect Identification}
\begin{itemize}[leftmargin=*,noitemsep]
  \item Locate and analyze the aspect in context
  \item Extract explicit descriptions related to the aspect
  \item \textit{Evidence focus:} Identify key phrases describing the aspect
\end{itemize}

{Stage 2: Opinion Inference}
\begin{itemize}[leftmargin=*,noitemsep]
  \item Deduce underlying attitudes toward the aspect
  \item Connect linguistic cues to real-world implications
  \item \textit{Interpretation focus:} Bridge literal meaning to sentiment implications
\end{itemize}

{Stage 3: Polarity Prediction}
\begin{itemize}[leftmargin=*,noitemsep]
  \item Predict sentiment polarity (positive / negative / neutral)
  \item Justify using language elements (modifiers, comparatives, etc.)
  \item Address any conflicting evidence
  \item \textit{Classification focus:} Synthesize evidence into final sentiment
\end{itemize}

Analyze the polarity of aspect terms in text: \texttt{POSITIVE, NEGATIVE, CONFLICT, NEUTRAL}.
}






\subsection{The Proposed CAPITAL}
In this subsection, we first formally define the variant of the front-door adjustment adopted for the ISA task, and then detail the procedure for estimating the causal effect of the input query $X$ on the polarity prediction $Y$.

As illustrated in Figure~\ref{fig:fig2c}, the variable $T$ satisfies all the conditions of the front-door criterion relative to $(X, Y)$. Therefore, $T$ can be used as a valid front-door adjustment variable to identify the causal effect of $X$ on $Y$. However, since intervention on $X$ is impossible in ISA with LLMs, we propose the following variant of the front-door adjustment:
\begin{equation}
\label{001}
    \begin{aligned}
        P(Y \mid \mathrm{do}(X)) = \sum_{T} P(T \mid \mathrm{do}(X)) \cdot P(Y \mid \mathrm{do}(T))
    \end{aligned}
\end{equation}

\subsubsection{Inference of CoT Distributions via Clustering} 
We first describe how to estimate the effect of $X$ on $T$, i.e., $P(T \mid \mathrm{do}(X))$. In our setting, $X$ refers to the full input prompt provided to the LLM, which includes a fixed set of in-context demonstrations and a test query. Since we can fully control and specify the content of prompt, intervening on $X$ by setting it to a particular prompt is feasible and aligns with the semantics of the do-operator in causal inference. By repeatedly prompting the LLM with the same $X$ and sampling the resulting CoT reasoning paths $T$, we can empirically approximate the interventional distribution $P(T \mid \mathrm{do}(X))$. 

To improve sampling diversity, we follow prior work~\cite{wangself} and adjust the temperature parameter of the LLM to generate multiple distinct CoTs. These CoTs are then embedded, clustered, and used to derive representative reasoning paths and their corresponding probabilities.

We select in-context demonstration $d$ from the training set based on their similarity to the test case. These demonstrations are then combined with the test sentence $s^{\text{test}}$ to form the final prompt. The structure of the final prompt $x$ is defined as:
\begin{equation}
    x = [d_1, \dots, d_r, s^{\text{test}}], \quad r = 1, \dots, R,
    \label{eq:test}
\end{equation}
where $r$ denotes the number of demonstration examples used in the few-shot prompting setup. Each $d$ consists of a demonstration sentence $s^{\text{demo}}$ and its corresponding demonstration CoT $t^{\text{demo}}$. 

Given the final prompt $x$, we allow the LLM to generate multiple distinct CoTs by adjusting its temperature parameter. This modification promotes greater output diversity, following a similar technique to that used in~\cite{zhang2025causal}. Consequently, the set of CoTs is obtained as follows:
\begin{equation}
    \{t_a \mid a = 1, \dots, A\} = \mathrm{LLM}(x),
\end{equation}where $a$ indexes each CoT, and $A$ denotes the total number of CoTs.

To conduct distance-based clustering, the generated CoTs $t_a$ are processed through an encoder to obtain their text embeddings $\bar{t_a}$. Following~\cite{devlin2019bert}, we augment the input with special tokens [CLS] and [SEP], and use the embedding of the [CLS] token as the representation of each CoT:
\begin{equation}
    \bar{t_a} = \mathrm{Encoder}([\mathrm{CLS}], t_a, [\mathrm{SEP}]).
\end{equation}

We then apply K-means clustering to the CoT embeddings to obtain $K$ clusters $T$:
\begin{equation}
    \{T_1, \dots, T_k\} = \mathrm{K\text{-}means}(\bar{t_1}, \dots, \bar{t_a}),
\end{equation}
where $T_k$ denotes the $k$-th cluster among the $K$ clusters.

According to the obtained clusters, we select representative CoTs $t^{\text{centroid}}_k$ by identifying the CoT nearest to each cluster's centroid:
\begin{equation}
    t^{\text{centroid}}_k = \mathrm{argmin}_{t \in T_k} \, \mathrm{dist}(t, \mu_k), \quad k = 1, \dots, K,
    \label{eq:cluster}
\end{equation}
where $\mu_k$ denotes the centroid of cluster $T_k$, and $\mathrm{dist}(\cdot, \cdot)$ is the distance function in the embedding space.

The embedding $\bar{t}^{\text{centroid}}_k$ is computed as follows:
\begin{align}
    \bar{t}^{\text{centroid}}_k = \text{Encoder}([\text{CLS}], t^{\text{centroid}}_k, [\text{SEP}])
\end{align}

The causal effect of $X$ on $T$ can thus be estimated by the relative size of each cluster:
\begin{equation}
\label{002}
    P(t^{\text{centroid}}_k \mid \mathrm{do}(X)) \approx \frac{|T_k|}{A},
\end{equation}
where $|T_k|$ denotes the number of CoTs assigned to the $k$-th cluster.

\subsubsection{Polarity Estimation via Iterative CoT} We now describe how to estimate $P(Y \mid \mathrm{do}(T))$, which quantifies the causal effect of the generated CoT on the final polarity prediction. Following Theorem~\ref{eq:backdoor}, the interventional distribution $P(Y \mid \mathrm{do}(T))$ can be estimated via the backdoor adjustment formula as follows:
\begin{align}
    P(Y \mid \mathrm{do}(T)) 
    &= \sum_{X} P(Y \mid T, X) P(X) \\
    &= \mathbb{E}_{X} \left[ P(Y \mid T, X) \right].
\end{align}

The equivalence between the summation form and the expectation form follows directly from the definition of expectation. Specifically, $\mathbb{E}_{X} [P(Y \mid T, X)]$ denotes the weighted average of $P(Y \mid T, X)$ over the distribution $P(X)$, which is precisely the meaning of the summation $\sum_{X} P(Y \mid T, X) P(X)$.

The value space of prompt is typically intractable to enumerate, and prior studies have adopted the normalised weighted geometric mean (NWGM) approximation to address this issue~\cite{tian2022debiasing,chen2023causal}. Inspired by~\cite{zhang2025causal}, we propose a prompt-based adaptation of the NWGM approximation for $\mathbb{E}_{X} [P(Y \mid T, X)]$ by integrating encoder-based intervention and in-context learning (ICL) prompting. The core idea of NWGM is to enhance the representation of the CoT $t_k$ with an auxiliary embedding that captures as much relevant sample information from the input space $X$ as possible. However, due to the limited context length of LLMs, it is infeasible to include all training samples in a single prompt. Instead, we select only the most relevant samples to optimise the current reasoning path.

Concretely, we first employ an encoder to derive the embedding vector $\bar{t}_k$ for the $k$-th CoT $t_k$. We then retrieve ICL demonstrations from the training set by measuring the similarity between $\bar{t}_k$ and the embeddings of candidate examples. These demonstrations serve as a proxy to approximate the expectation over $X$ in the backdoor formula. Finally, the demonstrations are ranked based on cosine similarity to $\bar{t}_k$, such that more relevant samples are assigned greater importance in the prompt construction.

Given a training set $\mathcal{D} = \{d_j = (s_j, t_j^{\text{wrong}}, t_j^{\text{correct}})\}_{j=1}^J$, where $s_j$ denotes the sentence of the $j$-th training example, and $t_j^{\text{wrong}}$ and $t_j^{\text{correct}}$ are its corresponding incorrect and correct CoTs, respectively. Here, $J$ denotes the total number of training samples. The embeddings $\bar{t}_j^{\text{wrong}}$ of incorrect CoTs are computed as follows:
\begin{align}
    \bar{t}_j^{\text{wrong}} = \text{Encoder}([\text{CLS}], t_j^{\text{wrong}}, [\text{SEP}]).
\end{align}

In our framework, we only consider the incorrect CoTs $t_j^{\text{wrong}}$ when computing embeddings for demonstration retrieval. This design choice is motivated by the observation that incorrect CoTs more effectively reflect the model’s uncertainty or failure modes in reasoning, making them informative anchors for identifying similar error patterns. By retrieving demonstrations based on semantically similar mistakes, we can guide the model to revise and improve the current test-time CoT. Moreover, excluding correct CoTs from the retrieval process avoids direct leakage of ideal reasoning paths, thereby maintaining a realistic setting where no ground-truth reasoning is assumed to be available during inference.

Prior research~\cite{margatina2023active} has demonstrated that employing demonstration examples semantically similar to test instances enhances ICL performance. Thus, the back-door intervention can be approximated by retrieving the most similar training examples based on the CoT embedding $\bar{t}_k^{\text{centroid}}$. Specifically, we rank the training set $\mathcal{D}$ in descending order of cosine similarity between $\bar{t}_k^{\text{centroid}}$ and $\bar{t}_j^{\text{wrong}}$:
\begin{equation}
    \{t_j^{\text{wrong} \uparrow}\}_{j=1}^{J} = \mathrm{Rank}(\mathcal{D}, \bar{t}_k^{\text{centroid}}, \{\bar{t}_j^{\text{wrong}}\}_{j=1}^J),
\end{equation}
where $t_j^{\text{wrong} \uparrow}$ denotes the $j$-th demonstration in the ranked list, and $\mathrm{Rank}$ indicates that the demonstrations are ordered such that $\cos(\bar{t}_k^{\text{centroid}}, \bar{t}_i^{\text{wrong}}) \ge \cos(\bar{t}_k^{\text{centroid}}, \bar{t}_j^{\text{wrong}})$ for $i < j$.

For each $t_k^{\text{centroid}}$, the final prompt is constructed as:
\begin{equation}
    x_{t_k^{\text{centroid}}} = [t_l^{\text{wrong} \uparrow}, \dots, t_1^{\text{wrong} \uparrow}, s^{\text{test}}], \quad l = 1, \dots, L
\end{equation} where $L$ denotes the number of top-ranked demonstration examples selected based on their similarity.

We then query the LLM $n$ times using the prompt along with the original CoT $t_k^{\text{centroid}}$, resulting in $n$ improved CoTs and corresponding polarity predictions:
\begin{equation}
    \{(y_{k, n}) \mid n = 1, \dots, N\} = \mathrm{LLM}(x_{t_k^{\text{centroid}}}, t_k^{\text{centroid}}).
\end{equation}

The probability of the predicted polarity can now be approximated as follows:
\begin{equation}
\label{003}
    \mathbb{E}_{x} \left[ P(Y \mid T, X) \right] \approx \frac{1}{N} \sum^{N}_{n=1} \mathbb{I}(y_{k, n} = y),
\end{equation}
where $\mathbb{I}(\cdot)$ is the indicator function.

Intuitively, this approximation treats the LLM as a black-box sampler: for a fixed reasoning path $t$, we construct multiple input prompts by varying the retrieved demonstrations. By querying the LLM $N$ times with these varying prompts, we obtain a set of output predictions $\{y_{k, n}\}$. The frequency with which a specific label $y$ appears reflects its empirical probability under the distribution $P(Y \mid \mathrm{do}(T))$.

\subsubsection{Final Output} Based on Equations~\ref{001}, \ref{002}, and \ref{003}, we obtain the final approximation of the causal effect of $X$ on $Y$ as follows:
\begin{equation}
    \begin{aligned}
        P(Y \mid \mathrm{do}(X)) \approx \sum^{K}_{k=1} \frac{|T_k|}{A} \cdot \frac{1}{N} \sum^{N}_{n=1} \mathbb{I}(y_{k, n} = y),
    \end{aligned}
\end{equation} where $T_k$ denotes the $k$-th cluster of reasoning paths (i.e., CoTs), and $|T_k|$ is the number of CoTs within the cluster. The constant $A$ represents the total number of generated CoTs before clustering. The term $\mathbb{I}(y_{k, r} = y)$ is an indicator function that returns 1 if the $r$-th prediction $y_{k, r}$ (generated from the prompt conditioned on the $k$-th cluster's centroid CoT) matches the target polarity label $y$, and 0 otherwise.

\section{Experiments}
\label{sec:experiments}
\subsection{Datasets}

We conduct experiments on the SemEval14 Laptop and Restaurant datasets~\cite{pontiki2016semeval}, where instances are categorised into explicit and implicit sentiment types following the annotation protocol of~\cite{li2021learning}. Both datasets consist of customer reviews annotated with aspect terms (e.g., "battery" or "service") and their associated sentiment polarities (positive, negative, neutral, or conflict). The Laptop dataset comprises electronic product reviews that often contain technical jargon and fragmented expressions. In contrast, the Restaurant dataset consists of restaurant reviews featuring more subjective and descriptive language. We evaluate all methods using the F1-score, which reflects the balance between precision and recall in sentiment classification.

\subsection{Baselines}
Our framework is compared with the following approaches to evaluate its effectiveness:
\begin{itemize}
    \item \textbf{In-context Learning (ICL)}~\cite{DBLP:conf/nips/BrownMRSKDNSSAA20} prompts LLMs using a few demonstration examples that include only questions and their corresponding answers, without any intermediate reasoning steps or explanations.
    
    \item \textbf{Chain-of-Thought (CoT)}~\cite{DBLP:conf/nips/Wei0SBIXCLZ22} supplies LLMs with demonstration examples containing detailed reasoning processes, guiding the model step-by-step to derive the correct answer.
    
    \item \textbf{CoT with Self-Consistency (CoT-SC)}~\cite{wangself} extends CoT prompting by generating multiple reasoning chains for a given query and selecting the most frequent answer through majority voting.
    
    \item \textbf{Context-aware Decoding (CAD)}~\cite{DBLP:conf/naacl/ShiHLTZY24} improves reasoning reliability by comparing LLM output distributions with and without additional contextual information during decoding.
    
    \item \textbf{THOR}~\cite{fei2023reasoning} adopts a three-hop prompting strategy combined with a voting mechanism to infer implicit sentiment polarity more effectively.
\end{itemize}

In our experiments, we adopt three pre-trained LLMs as backbone models to ensure diversity and comparability: LLaMA-2~\cite{DBLP:journals/corr/abs-2307-09288}, LLaMA-3~\cite{DBLP:journals/corr/abs-2402-16048}, and GPT-3.5 Turbo~\cite{BrownMRSKDNSSAA20}. These models differ in parameter scales, training paradigms, and openness (open-source vs. closed-source), providing a diverse and robust foundation for comprehensive evaluation.

\subsection{Implementation Details}
We conduct our experiments on a high-performance computing system equipped with an Intel Core i9-13900K CPU and an NVIDIA A6000 GPU (48GB VRAM). LLaMA-2 7B is deployed locally, while LLaMA-3 253B is accessed via NVIDIA’s hosted API~\footnote{\url{https://build.nvidia.com/nvidia/llama-3-1-nemotron-ultra-253b-v1}}. GPT-3.5 is interfaced through OpenAI’s official API~\footnote{\url{https://openai.com}}. We set key hyper-parameters as follows: the number of demonstrations $R=3$, cluster size $K=8$, number of CoT samples $A=20$, and second-stage prompting iterations $N=5$. All baseline models follow their respective optimal settings as reported in the original publications.

\subsection{Main Results}
Table~\ref{tab:main} presents a comprehensive comparison of the proposed framework {CAPITAL} with five competitive baselines: ICL, CoT, CoT-SC, CAD, and THOR, across three backbone LLMs: LLaMA-2, LLaMA-3, and GPT-3.5. The evaluation includes two settings: ALL, which covers both ESA and ISA, and ISA, which focuses solely on the implicit component. Across all models, a consistent performance progression is observed: ICL performs the worst, followed by CoT and then CoT-SC. This trend confirms that incorporating step-by-step reasoning and self-consistency sampling improves LLM performance, which aligns with previous research findings~\cite{BrownMRSKDNSSAA20, DBLP:conf/nips/Wei0SBIXCLZ22, wangself}. CAD and THOR contribute additional enhancements by introducing external context and multi-hop prompting, but the gains from these approaches remain moderate when compared to CAPITAL.

Our framework achieves the best results across all configurations, significantly outperforming existing baselines on both datasets and across all backbone models. For instance, under the LLaMA-2 setting, it reaches an F1-score of 62.37 on the ISA subset of the Restaurant dataset, exceeding THOR by more than 7 points. With LLaMA-3, the method attains scores of 71.63 on ISA (Restaurant) and 73.68 on ISA (Laptop), surpassing the next-best approach by at least 4 points. Even when using the more constrained GPT-3.5 model, performance improves from 54.33 (THOR) to 67.64, representing a substantial relative gain. These results demonstrate that by estimating and leveraging causal effects, our approach produces more robust and reliable predictions, particularly for complex reasoning tasks such as ISA. On average, it achieves a 4\% to 10\% improvement over the strongest competing method, underscoring its practical effectiveness in handling implicit sentiment.

\subsection{Robustness Analysis}
Existing causality-based prompting methods~\cite{zhang2025causal} commonly utilize out-of-distribution (OOD) datasets to evaluate a model’s robustness to spurious correlations and latent biases. Following this protocol, we evaluate our proposed approach on both original (in-distribution) and adversarially perturbed (OOD) versions of the SemEval datasets. The adversarial sets are constructed by subtly modifying input samples to preserve semantics while introducing confounding structures that often mislead non-causal models. We focus this robustness study on LLaMA-3, the strongest backbone in our experiments.

Table~\ref{tab:Rob} presents a detailed comparison of performance across methods. All baselines exhibit noticeable performance degradation under adversarial conditions, confirming the vulnerability of current prompting strategies to distributional shifts. In contrast, our framework consistently achieves the highest F1-scores across both domains and tasks. Specifically, on the Restaurant dataset, it reaches 94.59 on the original set and 90.17 on the adversarial version. For the Laptop dataset, the scores are 87.09 and 83.56, respectively. These results suggest that by explicitly modeling causal relationships through front-door adjustment, our method maintains greater predictive stability and generalises more effectively to challenging, confounded scenarios.

\begin{table}[t]
  \centering
    \caption{Comparison F1-scores of CAPITAL and five methods across three backbone LLMs on two ISA tasks. The best results are highlighted in bold.}
  \label{tab:main}
  \begin{tabular}{c|c|cc|cc}
    \toprule
    \multirow{2}{*}{\textbf{Model}} & \multirow{2}{*}{\textbf{Method}} & \multicolumn{2}{c|}{\textbf{Restaurant}} & \multicolumn{2}{c}{\textbf{Laptop}} \\ 
     & & ALL~$\uparrow$ & ISA~$\uparrow$ & ALL~$\uparrow$ & ISA~$\uparrow$ \\
    \midrule
    \multirow{7}{*}{LLaMA-2} 
        & ICL         & 63.45 & 44.61 & 56.35 & 41.72 \\
        & CoT         & 70.44 & 53.87 & 63.11 & 47.56 \\
        & CoT-SC      & 73.05 & 55.67 & 64.76 & 50.73 \\
        & CAD         & 72.85 & 54.61 & 62.15 & 52.15 \\
        & THOR        & 74.36 & 55.19 & 63.13 & 49.23 \\ 
        & CAPITAL  & \textbf{80.31} & \textbf{62.37} & \textbf{75.03} & \textbf{57.25} \\
    \midrule
    \multirow{7}{*}{LLaMA-3} 
        & ICL         & 77.65 & 55.39 & 70.30 & 53.45 \\
        & CoT         & 85.39 & 63.30 & 77.61 & 62.86 \\
        & CoT-SC      & 90.65 & 67.79 & 79.14 & 65.71\\
        & CAD         & 87.31 & 63.51 & 76.33 & 62.01 \\
        & THOR        & 88.12 & 66.31 & 78.51 & 63.55 \\
        & CAPITAL  & \textbf{94.59} & \textbf{71.63} & \textbf{87.09} & \textbf{73.68} \\
    \midrule
    \multirow{7}{*}{GPT-3.5} 
        & ICL         & 71.22 & 48.63 & 60.95 & 45.31 \\
        & CoT         & 77.61 & 51.40 & 66.38 & 49.14 \\
        & CoT-SC      & 79.43 & 51.84 & 67.27 & 52.15 \\
        & CAD         & 78.31 & 50.16 & 66.58 & 48.32 \\
        & THOR        & 80.15 & 54.33 & 68.91 & 55.17 \\
        & CAPITAL  & \textbf{90.67} & \textbf{67.64} & \textbf{79.78} & \textbf{65.47} \\
    \midrule
    \end{tabular}
\end{table}

\begin{table}[t]
  \centering
    \caption{Robustness evaluation results on the LLaMA-3 model. {Ori.} denotes the original (in-distribution) dataset, and {Adv.} refers to the adversarial (out-of-distribution) version. The best results for each setting are highlighted in {bold}.}
  \label{tab:Rob}
  \begin{tabular}{c|cc|cc}
    \toprule
    \multirow{2}{*}{\textbf{Method}} & \multicolumn{2}{c|}{\textbf{Restaurant}} & \multicolumn{2}{c}{\textbf{Laptop}} \\ 
     & Ori. & Adv. & Ori. & Adv. \\
    \midrule
        ICL         & 77.65 & 68.37 & 70.30 & 58.47 \\
        CoT         & 85.39 & 74.07 & 77.61 & 65.44 \\
        CoT-SC      & 90.65 & 81.65 & 79.14 & 68.37 \\
        CAD         & 87.31 & 79.55 & 76.33 & 69.59 \\
        THOR        & 88.12 & 82.14 & 78.51 & 74.33 \\ 
        CAPITAL        & \textbf{94.59} & \textbf{90.17} & \textbf{87.09} & \textbf{83.56} \\
    \midrule
  \end{tabular}
\end{table}

\subsection{Impact of Hyper-parameters}

\begin{figure}[t]
	\centering
	\begin{subfigure}[b]{0.45\linewidth}
		\centering
		\includegraphics[width=\linewidth]{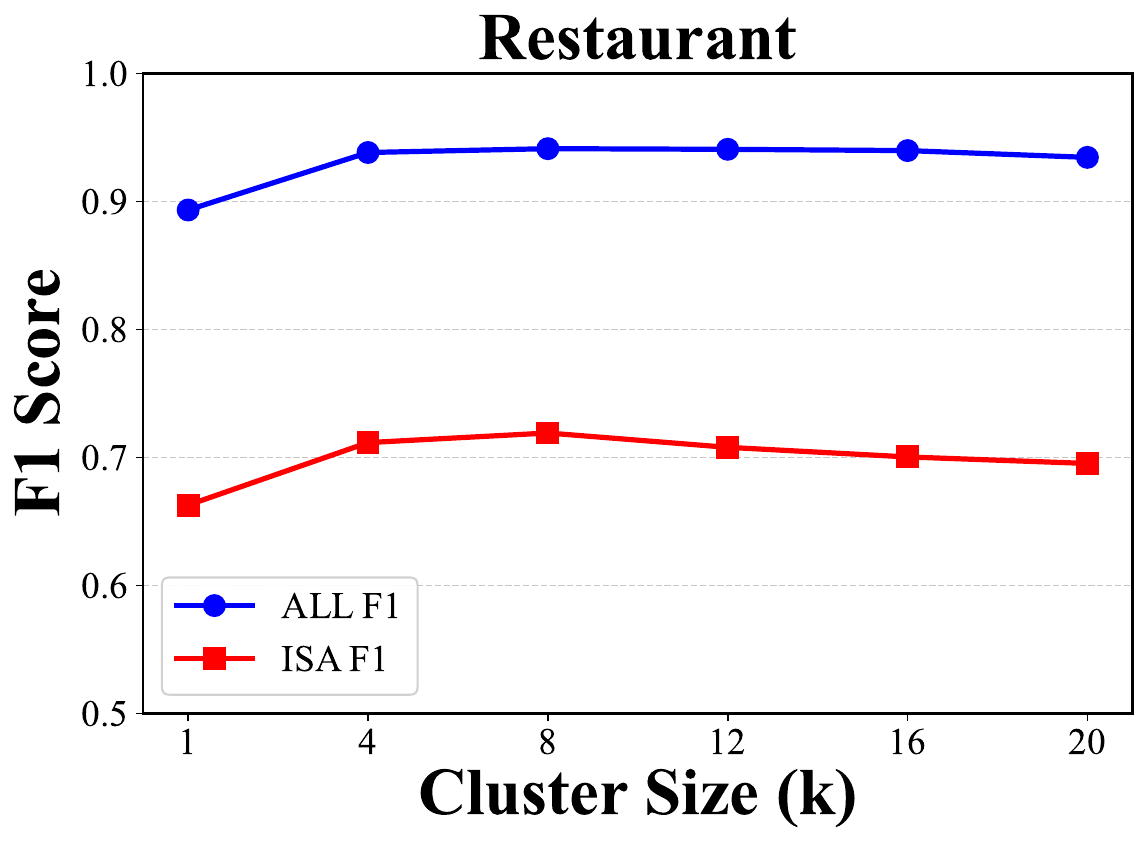}
	\end{subfigure}
	\hfill
	\begin{subfigure}[b]{0.45\linewidth}
		\centering
		\includegraphics[width=\linewidth]{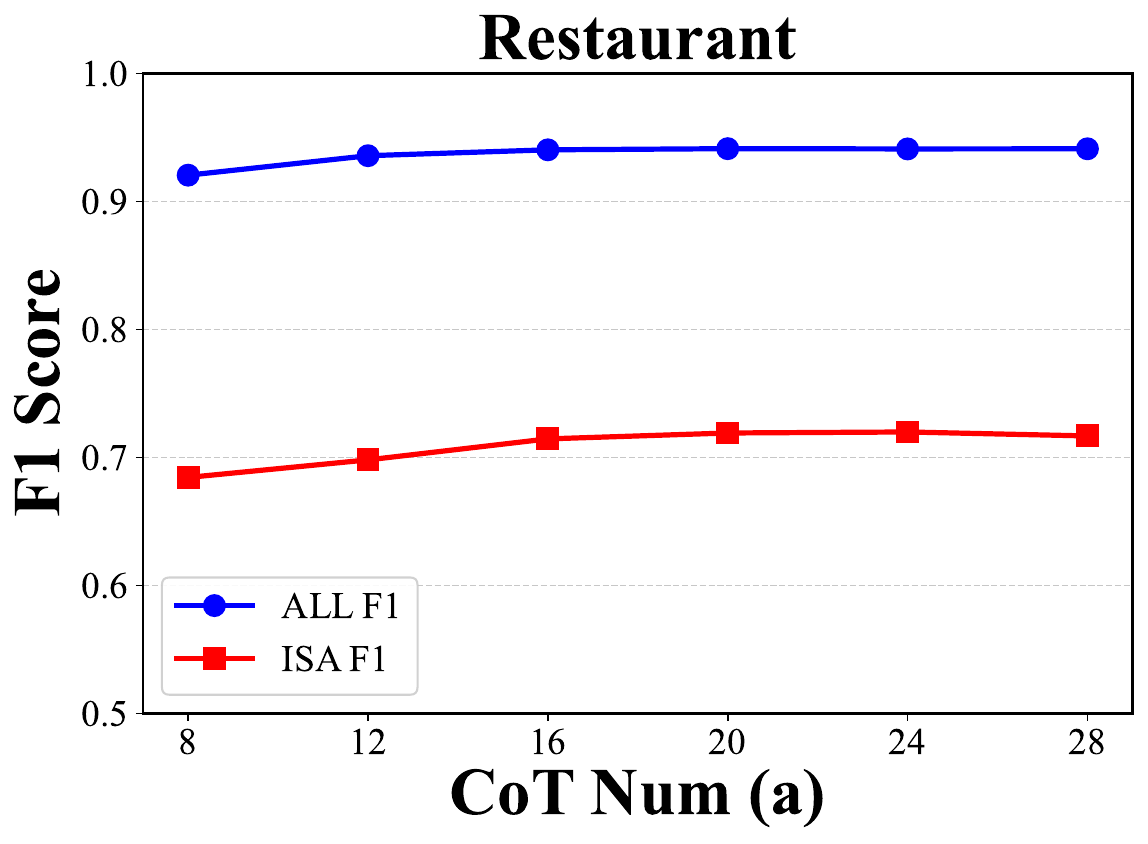}
	\end{subfigure}
	\vskip\baselineskip
	\begin{subfigure}[b]{0.45\linewidth}
		\centering
		\includegraphics[width=\linewidth]{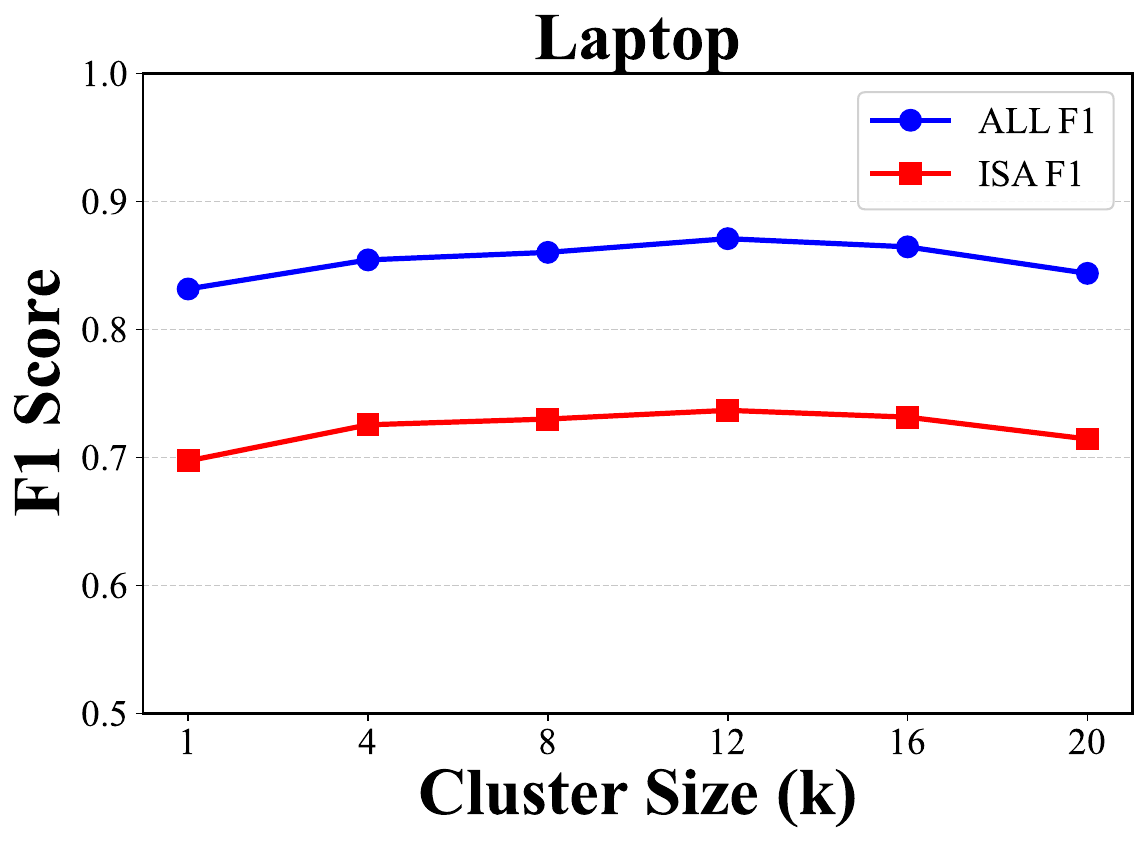}
	\end{subfigure}
	\hfill
	\begin{subfigure}[b]{0.45\linewidth}
		\centering
		\includegraphics[width=\linewidth]{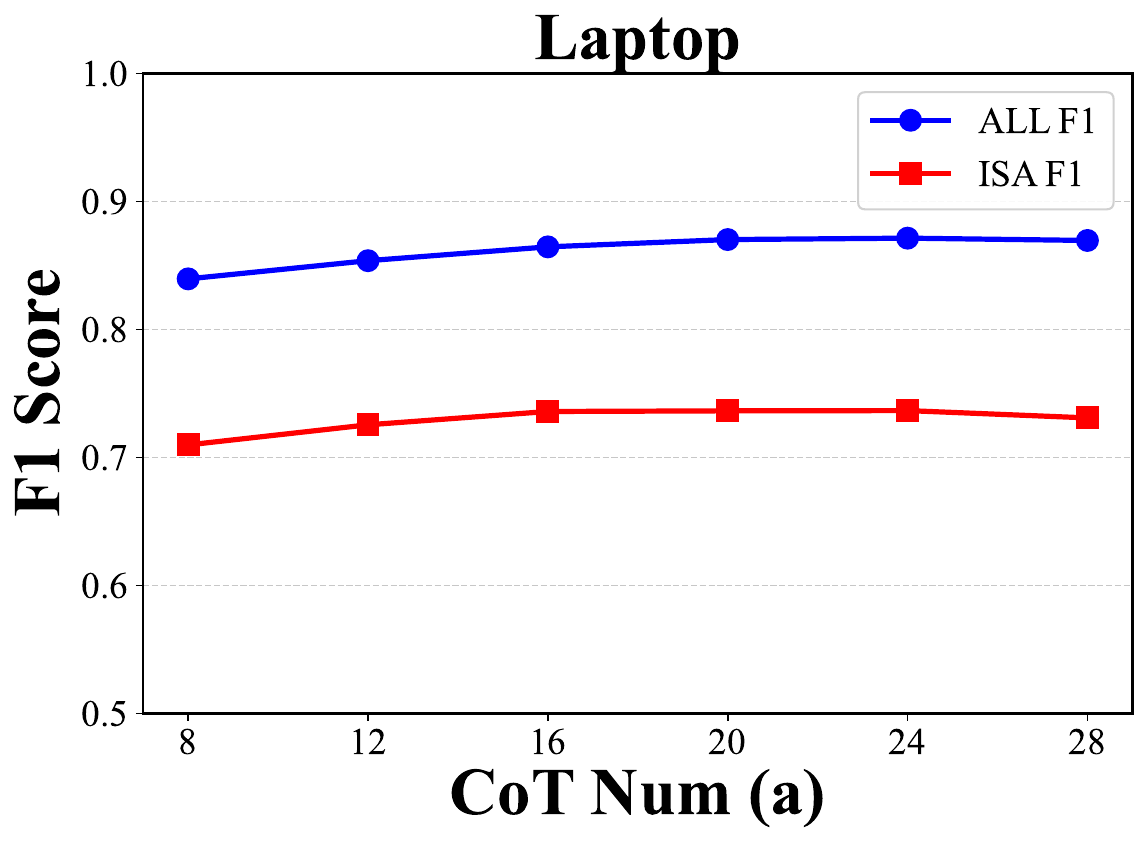}
	\end{subfigure}
	\caption{Hyper-parameter analysis of the number of clusters and the number of CoTs on framework performance.}
	\label{fig:hyper}
\end{figure}

We conduct additional hyper-parameter studies to examine how the number of clusters  and the number of CoTs influence model performance. As illustrated in Figure~\ref{fig:hyper}, performance on both implicit and explicit sentiment analysis tasks increases significantly as the number of clusters grows from 1 to 8 on the Restaurant dataset and from 1 to 12 on the Laptop dataset. Beyond these values (i.e., $K > 8$ for Restaurant and $K > 12$ for Laptop), the performance either stabilises or slightly declines. This observation suggests that when the number of clusters is too small, the CoT distribution is insufficiently captured, whereas an excessively large $K$ may dilute the information within each cluster, leading to unreliable estimation of $P(Y \mid \mathrm{do}(X))$ due to limited CoT samples. Therefore, we set $K=8$ for the Restaurant dataset and $K=12$ for the Laptop dataset to balance estimation quality and computational cost. Regarding the number of CoTs $A$, both datasets exhibit consistent performance improvements as $A$ increases from 8 to 20. However, further increases beyond 20 result in marginal or negative gains. Thus, we fix $A=20$ in our experiments, as it provides a sufficient trade-off between performance and computational efficiency.

\subsection{Ablation Study}
As shown in Table~\ref{tab:Abl}, we conduct an ablation study to investigate the contributions of three key components in our framework: (1) the NWGM approximation strategy, (2) the K-means clustering module, and (3) the final-stage weighting mechanism. The analysis is performed on both the Restaurant and Laptop datasets using the LLaMA-3 model.

\subsubsection{NWGM Approximation}
To assess the impact of the NWGM approximation, we compare our method against two variants: \textit{NWGM-Reverse} and \textit{NWGM-Random}. In NWGM-Reverse, the order of in-context demonstrations is reversed, weakening the proximity-based influence of high-similarity examples. In NWGM-Random, demonstrations are randomly selected from the training set without any similarity-based filtering. Both variants result in consistent performance drops, especially in the ISA subset. This confirms that ordering demonstrations by semantic similarity is crucial for the NWGM approximation to effectively estimate $P(Y \mid \mathrm{do}(T))$.

\subsubsection{K-means Clustering}
To evaluate the role of K-means clustering in estimating $P(T \mid \mathrm{do}(X))$, we remove the clustering step and instead randomly select $K$ CoTs in the first stage of front-door adjustment, assigning them equal weights ($1/K$). This variant, denoted as \textit{w/o K-means}, leads to noticeable performance degradation, particularly in the ISA task. These results suggest that clustering helps identify representative reasoning patterns, which are essential for reliable estimation of intermediate causal effects.

\subsubsection{Weighting Mechanism}
Finally, we investigate the importance of the weighting mechanism by replacing our weighted aggregation with simple majority voting over final answers. This variant is referred to as \textit{w/o Weighting}. The results show a substantial decline in accuracy, especially on the ISA subset of the Laptop dataset (from 83.56 to 78.94). This validates the effectiveness of computing the final output as a joint product of $P(T \mid \mathrm{do}(X))$ and $P(Y \mid \mathrm{do}(T))$, rather than relying on frequency-based decision heuristics.

Overall, the ablation study demonstrates that each component in our framework plays a vital role in boosting the robustness and precision of causal reasoning. The full configuration consistently achieves the best results across all settings.

\begin{table}[!t]
  \centering
  \caption{The results of ablation study on LLaMA3. The best results are in bold.}
  \label{tab:Abl}
  \begin{tabular}{l|cc|cc}
    \toprule
    \multirow{2}{*}{\textbf{Method}} & \multicolumn{2}{c|}{\textbf{Restaurant}} & \multicolumn{2}{c}{\textbf{Laptop}} \\ 
     & ALL & ISA & ALL & ISA \\
    \midrule
        CAPITAL              & \textbf{94.59} & \textbf{90.17} & \textbf{87.09} & \textbf{83.56} \\
        \quad NWGM-Reverse         & 94.01 & 89.37 & 86.33 & 82.84 \\
        \quad NWGM-Random          & 93.48 & 88.76 & 85.72 & 81.93 \\
        \quad w/o K-means          & 92.64 & 88.15 & 85.81 & 80.17 \\
        \quad w/o Weighting        & 92.37 & 87.59 & 85.04 & 78.94 \\ 
    \midrule
  \end{tabular}
\end{table}

\section{Conclusion}
\label{sec:conclusion}s
This paper presents CAPITAL, a causal prompting framework for implicit sentiment analysis that incorporates front-door adjustment into chain-of-thought reasoning. To operationalise this framework, we decompose the overall causal effect into two components: the influence of the input prompt on the reasoning chains and the influence of these chains on the final prediction. CAPITAL estimates these components using encoder-based clustering and the NWGM approximation. To enhance estimation accuracy, a contrastive learning strategy is used to fine-tune the encoder so that its representations align more closely with the latent reasoning space of the LLM. Experimental results on three large language models and two benchmark datasets show that the proposed method leads to substantial improvements in both explicit and implicit sentiment prediction. CAPITAL consistently outperforms strong prompting baselines and demonstrates greater robustness on adversarial data. Ablation and hyper-parameter studies further validate the effectiveness of each component. This work offers a principled approach to incorporating causal inference into prompting and suggests promising directions for bias-aware reasoning in large language models. Beyond sentiment analysis, the proposed causal effect estimation strategy offers a generalisable method applicable to a wide range of reasoning-intensive NLP tasks.

\bibliographystyle{IEEEtran}
\bibliography{mybib}


\newpage

\vspace{11pt}


\vspace{11pt}

\vfill

\end{document}